%% file: ijcai24.tex
\documentclass{article}
\usepackage{ijcai24}

\usepackage{times}
\usepackage{soul}
\usepackage{url}
\usepackage[hidelinks]{hyperref}
\usepackage[utf8]{inputenc}
\usepackage[small]{caption}
\usepackage{graphicx}
\usepackage{amsmath}
\usepackage{amsthm}
\usepackage{booktabs}
\usepackage{algorithm}
\usepackage{algorithmic}
\usepackage[switch]{lineno}
\usepackage{booktabs}
\usepackage{fontawesome}
\usepackage{tikz}
\usepackage{inconsolata}
\usepackage{pifont}
\usepackage[edges]{forest}
\usepackage{makecell}
\usepackage{cleveref}
\usepackage{setspace}
\usepackage[utf8]{inputenc}
\definecolor{hidden-draw}{RGB}{205, 44, 36}
\definecolor{hidden-blue}{RGB}{194,232,247}
\definecolor{hidden-orange}{RGB}{243,202,120}
\definecolor{hidden-yellow}{RGB}{255,229,204}
\definecolor{hidden-red}{RGB}{255,204,204}
\definecolor{hidden-draw}{RGB}{20,68,106}
\definecolor{hidden-pink}{RGB}{255,245,247}

\title{Continual Learning for Large Language Models: A Survey}

\author{
Tongtong Wu$^1$\and
Linhao Luo$^1$\and
Yuan-Fang Li$^{1}$\and
Shirui Pan$^2$\and
Thuy-Trang Vu$^1$\and \\
Gholamreza Haffari$^1$
\affiliations
$^1$Monash University{ }
$^2$Griffith University\\
\emails
\{first-name.last-name\}@monash.edu,
s.pan@griffith.edu.au
}

\begin{document}

\maketitle

\begin{abstract}
Large language models (LLMs) are not amenable to frequent re-training, due to high training costs arising from their massive scale. 
However, updates are necessary to endow LLMs with new skills and keep them up-to-date with rapidly evolving human knowledge. 
This paper surveys recent works on continual learning for LLMs. Due to the unique nature of LLMs, we catalog continue learning techniques in a novel multi-staged categorization scheme, involving continual pretraining, instruction tuning, and alignment. We contrast continual learning for LLMs with simpler adaptation methods used in smaller models, as well as with other enhancement strategies like retrieval-augmented generation and model editing. Moreover, informed by a discussion of benchmarks and evaluation, we identify a number of challenges and future work directions for this crucial task. 
% This paper emphasizes the necessity of frequent updates in large language models (LLMs) to ensure their alignment with the rapidly evolving landscape of human knowledge and language. It distinguishes continual learning from other enhancement strategies like retrieval-augmented generation and model editing, underscoring its focus on advancing LLMs' overall linguistic and cognitive capabilities. The paper highlights the unique multi-staged approach of continual learning in LLMs, involving continual pretraining, instruction tuning, and alignment, and contrasts this with the simpler adaptation methods used in smaller models. Setting itself apart, this study is the first to specifically address continual learning in LLMs, categorizing the discussion by the types of information updated and the stages of learning. The goal is to provide a comprehensive perspective on the effective implementation of continual learning in LLMs, paving the way for the development of more advanced and adaptable language models.
\end{abstract}

\input{sections/1_Intro}

\input{sections/2_Concept}

\input{sections/3_Pretraining}

\input{sections/4_Instruction}

\input{sections/5_Preference}

% \input{sections/6 Editting}
\input{sections/6_Evaluation}

\input{sections/7_Challenges_and_Future_Directions}

\input{sections/8_Conclusion}

\let\oldthebibliography\thebibliography
\let\endoldthebibliography\endthebibliography
\renewenvironment{thebibliography}[1]{
  \begin{oldthebibliography}{#1}
    \setlength{\itemsep}{0em}
    \setlength{\parskip}{0em}
}
{
  \end{oldthebibliography}
}
%% The file named.bst is a bibliography style file for BibTeX 0.99c
{
\setstretch{0.96}
\bibliographystyle{named}
\bibliography{ijcai24}
}
\end{document}

%% file: sections/1_Intro.tex
\section{Introduction}
% motivation
Recent years have witnessed the rapid advances of large language models' (LLMs) capabilities in solving a diverse range of problems. At the same time, it is vital for LLMs to be regularly updated to accurately reflect the ever-evolving human knowledge, values and linguistic patterns, calling for the investigation of \emph{continual learning} for LLMs. 
% In the dynamic landscape of artificial intelligence, the constant evolution and enhancement of large language models (LLMs) are crucial. The primary motivation for regularly updating these models lies in their need to accurately reflect the ever-changing human knowledge and linguistic patterns. Regular updates are vital to ensure that LLMs stay current with the latest trends, terminologies, and societal changes. 
% This paper explores the importance of updating LLMs and the methodologies employed, emphasizing the role of continual learning in maintaining the efficacy and relevance of these models in an environment where information is constantly evolving.
% continual learning v.s. RAG or Model Editing
Whilst continual learning bears some resemblance to other strategies for model improvements, such as retrieval-augmented generation (RAG)
~\cite{LewisPPPKGKLYR020} and model editing ~\cite{yao-etal-2023-editing}, their main purposes differ (\Cref{tab:comp}). 
Unlike  these strategies, whose primarily focus is on refining the domain-specific accuracy or expanding the model's factual knowledge base, continual learning aims to enhance the overall linguistic and reasoning capabilities of LLMs. This distinction is crucial as it shifts the focus from merely updating information to developing a model's ability to process and generate language in a more comprehensive and nuanced manner~\cite{ZhangFCNW23}.

\begin{figure}[t]
    \centering
    \includegraphics[width=.48\textwidth]{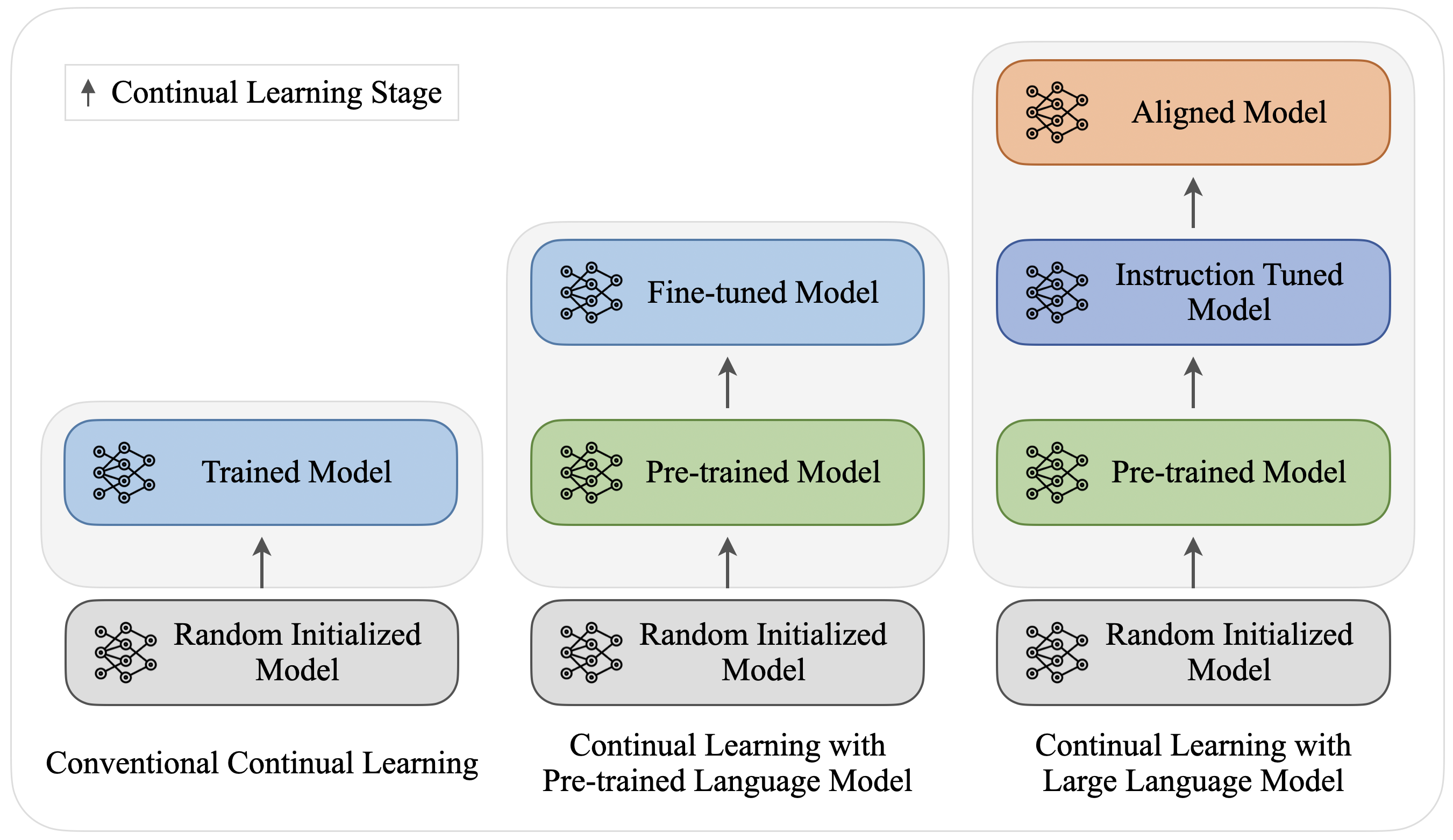}
    \caption{
    Continual learning for large language models involves hybrid multi-stage training with multiple training objectives.}
    %Continual learning for large language models involves multi-stage training, including pretraining, instruction learning and alignment, to incrementally incorporate different types of information.}
    \label{fig:cl4llm}
\end{figure}

% Our discussion includes a comparison between continual learning (CL) and other strategies for model improvement, such as retrieval-augmented generation (RAG)
% ~\cite{LewisPPPKGKLYR020} 
% %~\cite{LewisPPPKGKLYR020,abs-2312-10997} 
% and model editing ~\cite{yao-etal-2023-editing}. 
% % ~\cite{yao-etal-2023-editing,abs-2401-01286,abs-2401-01286}. 
% Unlike these strategies, whose primarily focus on expanding the model’s factual knowledge base or refining its domain-specific accuracy, continual learning aims to enhance the overall linguistic and reasoning capabilities of LLMs. This distinction is crucial as it shifts the focus from merely updating information to developing a model’s ability to process and generate language in a more comprehensive and nuanced manner~\cite{ZhangFCNW23}.
%~\cite{abs-2311-05876,ZhangFCNW23}.

\begin{figure*}[tb]
    \centering
    \includegraphics[width=\textwidth]{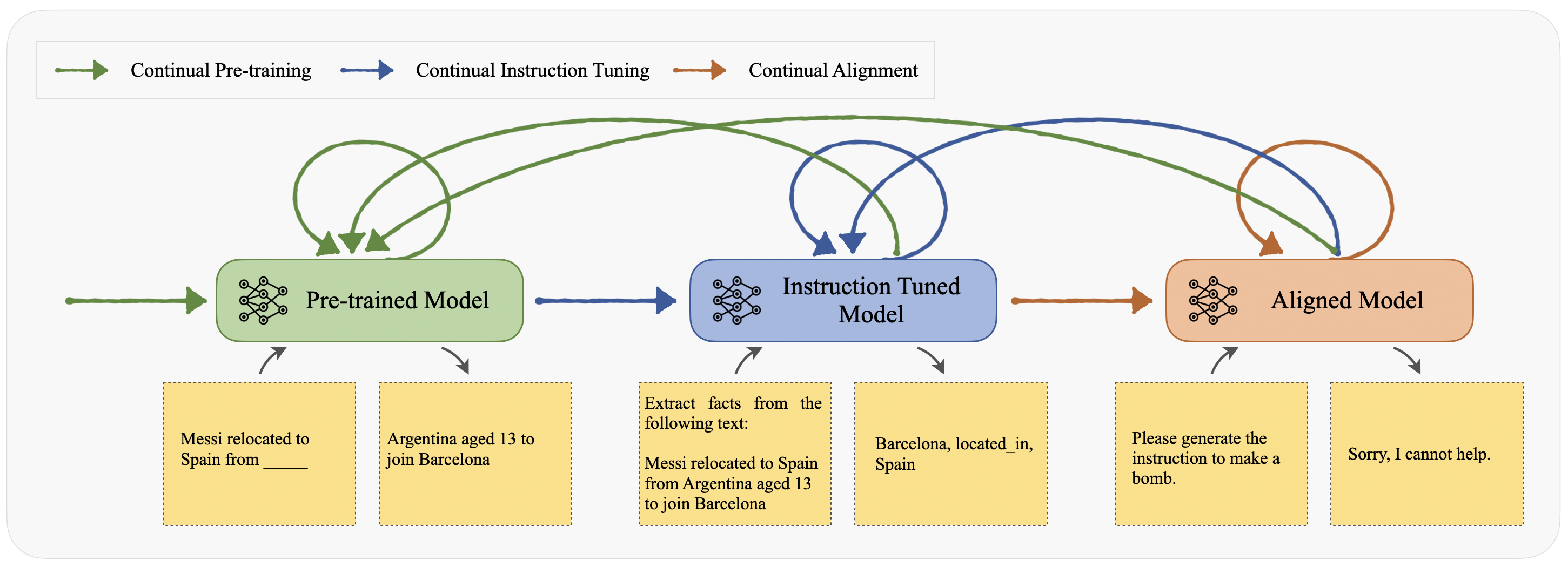}
    \caption{The continual learning of LLMs involves multi-stage and cross-stage iteration, which may lead to substantial forgetting problems. For example, when the instruction-tuned model resumes continual pre-training, it may encounter cross-stage forgetting, resulting in reduced performance on instruction-following tasks.}
    \label{fig:stage}
\end{figure*}

% continual learning for LLM v.s. smaller model
Continual learning for LLMs also differs from its use in smaller models, including smaller pre-trained language models (PLMs). Due to their vast size and complexity, LLMs require a multi-faceted approach to continual learning. We categorise it into three different stages, i.e.\ \emph{continual pretraining} to expand the model's fundamental understanding of language~\cite{JinZZ00WA022},
%~\cite{JinZZ00WA022,abs-2308-04014},
\emph{continual instruction tuning} to improve the model's response to specific user commands~\cite{zhang2023citb}, and \emph{continual alignment} to ensure the model's outputs adhere to values, ethical standards and societal norms~\cite{zhang2023copf}. This multi-stage process is distinct from the more linear adaptation strategies used in smaller models, as illustrated in \Cref{fig:cl4llm}, highlighting the unique challenges and requirements of applying continual learning to LLMs.

% This survey v.s. the previous surveys
This survey differentiates itself from previous studies by its unique focus and structure. While previous surveys in the field are typically organized around various continual learning strategies~\cite{biesialska-etal-2020-continual}, ours is the first to specifically address continual learning in the context of LLMs. We structure our analysis around the types of information that is updated continually and the distinct stages of learning involved in LLMs. This survey offers a detailed and novel perspective on how continual learning is applied to LLMs, shedding light on the specific challenges and opportunities of this application. Our goal is to provide a thorough understanding of the effective implementation of continual learning in LLMs, contributing to the development of more advanced and adaptable language models in the future.

\begin{table}[t]
\centering
\resizebox{\columnwidth}{!}{
\begin{tabular}{@{}ccc|c@{}}
\toprule
Information       & RAG & Model Editing & Continual Learning \\ \midrule
Fact        & \faCheckCircleO   & \faCheckCircleO             & \faCheckCircleO                  \\
Domain            & \faCheckCircleO   & ×             & \faCheckCircleO                  \\
Language          & ×   & ×             & \faCheckCircleO                  \\
Task              & ×   & ×             & \faCheckCircleO                  \\
Skills (Tool use) & ×   & ×             & \faCheckCircleO                  \\
Values            & ×   & ×             & \faCheckCircleO                  \\
Preference        & ×   & ×             & \faCheckCircleO                  \\ \bottomrule
\end{tabular}}
\caption{Continual Learning v.s.\ RAG and Model Editing}\label{tab:comp}
\end{table}

%% file: sections/2_Concept.tex
\section{Preliminary and Categorization}
\subsection{Large Language Model}
% - Introduce the architecture and the multiple training stages of LLMs.
Large language models (LLMs) like ChatGPT\footnote{\url{https://openai.com/blog/chatgpt}} and LLaMa \cite{touvron2023llama}
%, and Mistral \cite{jiang2023mistral} 
have shown superior performance in many tasks.
%Most existing LLMs derive from the Transformer architecture \cite{vaswani2017attention}, which consists of a stack of self-attention layers. 
They are usually trained in multiple stages, including pre-training, instruction tuning, and alignment, as illustrated in Figure \ref {fig:cl4llm}. In the \emph{pre-training} stage, LLMs are trained on a large corpus in a self-supervised manner \cite{dong2019unified}, where the training text is randomly masked and the LLMs are asked to predict the masked tokens. In the \emph{instruction tuning} stage, LLMs are fine-tuned on a set of instruction-output pairs in a supervised fashion \cite{zhang2023instruction}. Given a task-specific instruction as input, LLMs are asked to generate the corresponding output. In the \emph{alignment} stage, LLMs are further finetuned with human feedback to align their outputs with human expectations \cite{wang2023aligning}. The output of LLMs is scored by human annotators, and the LLMs are updated to generate more human-like responses.

\subsection{Continual Learning}
Continual learning focuses on developing learning algorithms to accumulate knowledge on non-stationary data, often delineated by classes, tasks, domains or instances. In supervised continual learning, a sequence of tasks $\left\{\mathcal{D}_1, \ldots, \mathcal{D}_T\right\}$ arrive in a streaming fashion. Each task $\mathcal{D}_t=\left\{\left(\boldsymbol{x}_i^t, y_i^t\right)\right\}_{i=1}^{n_t}$ contains a separate target dataset, where $\boldsymbol{x}_i^t\in \mathcal{X}_t$ , $\boldsymbol{y}_i^t\in \mathcal{Y}_t$. 
A single model needs to adapt to them sequentially, with only access to $\mathcal{D}_t$ at the t-th task. This setting requires models to acquire, update, accumulate, and exploit knowledge throughout their lifetime~\cite{biesialska-etal-2020-continual}. %\cite{wang2023comprehensive}. 

The major challenge conventional continuous learning tackles is that of \emph{catastrophic forgetting}, where the performance of a model on old tasks significantly diminishes when trained with new data. Existing studies can be roughly grouped into three categories, e.g., experience replay methods \cite{chaudhry2019tiny,WuLLHQZX21}, regularization-based methods \cite{kirkpatrick2017overcoming,0002GWQLD23}, and dynamic architecture methods \cite{mallya2018piggyback}. Recently, researchers have designed some hybrid methods that take advantage of the aforementioned techniques \cite{abs-2305-08698,he2024lifelong}. 

% \paragraph{Setup}
% - Class-incremental

% - Task-incremental

% - Domain-incremental

\subsection{Continual Learning for LLMs}
Continual Learning for Large Language Models aims to enable LLMs to learn from a continuous data stream over time. %, which is critical for LLMs to capture the ever-changing world. 
Despite the importance, it is non-trivial to directly apply existing continual learning settings for LLMs. We now  provide a forward-looking framework of continual learning for LLMs, then present a categorization of research in this area.

\paragraph{Framework}
Our framework of continual learning for LLMs is illustrated in Figure \ref{fig:stage}. W align continual learning for LLMs with the different training stages, including Continual Pre-training (CPT), Continual Instruction Tuning (CIT), and Continual Alignment (CA). The \emph{Continual Pre-training} stage aims to conduct training on a sequence of corpus self-supervisedly to enrich LLMs' knowledge and adapt to new domains. The \emph{Continual Instruction Tuning} stage finetunes LLMs on a stream of supervised instruction-following data, aiming to empower LLMs to follow users' instructions while transferring acquired knowledge for subsequent tasks. Responding to the evolving nature of human values and preferences, \emph{Continual Alignment (CA)} tries to continuously align  LLMs with human values over time.

While continual learning on LLMs can be conducted in each stage sequentially, the iterative application of continual learning also makes it essential to transfer across stages without forgetting the ability and knowledge learned from previous stages. For instance, we can conduct continual pre-training based on either instruction-tuned models or aligned models. However, we do not want the LLM to lose their ability to follow users' instructions and align with human values. Therefore, as shown in Figure \ref{fig:stage}, we use arrows with different colors to show the transfer between stages.

\paragraph{Categorization}
To better understand the research in this area, we provide a fine-grained categorization for each stage of the framework.

\textbf{Continual Pre-training (CPT)}
\begin{itemize}
    \item \emph{CPT for Updating Facts} includes works that adapt LLMs to learn new factual knowledge.
    \item \emph{CPT for Updating Domains} includes research that tailors LLMs to specific fields like medical and legal domains.
    \item \emph{CPT for Language Expansion} includes studies that extend the languages LLMs supports.
\end{itemize}

\textbf{Continual Instruction Tuning (CIT)}
\begin{itemize}
    \item \emph{Task-incremental CIT} contains works that finetune LLMs on a series of tasks and acquire the ability to solve new tasks.
    \item \emph{Domain-incremental CIT} contains methods that finetune LLMs on a stream of instructions to solve domain-specific tasks.
    \item \emph{Tool-incremental CIT} contains research that continually teaches LLMs to use new tools to solve problems.
\end{itemize}

\textbf{Continual Alignment (CA)}
\begin{itemize}
    \item \emph{Continual Value Alignment} incorporates studies that continually align LLMs with new ethical guidelines and social norms.
    \item \emph{Continual Preference Alignment} incorporates works that adapt LLMs to dynamically match different human preferences.
\end{itemize}

Besides categorizing methods based on training stages, we also provide an alternative categorization based on the information updated during continual learning. In Table \ref{tab:information}, we list some representative information that is updated for LLMs, e.g., facts, domains, tasks, values, and preferences. Based on the training objectives of LLMs, this information can be updated in different stages of LLM continual learning. 
The taxonomy in \Cref{fig:survey} shows our categorization scheme and recent representative work in each category. 

% Traditional continual learning involves various incremental setups like class-incremental, task-incremental, domain-incremental, and instance-incremental learning. However, in the context of LLMs, the concept of task boundaries can be blurred. Because LLMs are expected to be Artificial General Intelligence (AGI), which exhibits potential in many complex tasks.

% Instead, the general continual learning setup focuses on observing incremental data in an online manner without strict task distinctions, highlighting the adaptability and flexibility of these models in handling diverse tasks seamlessly.

% \paragraph{Strategies}
% - Introduce the basic concept of continual learning, including setup, and traditional continual learning methods.

% - Experience Replay Method

% - Regularization-based Method

% - Dynamic Architecture Method

\input{sections/texonomy}

% knowledge base ; planner/reasoner ; assistant? generative model / generation system

% how to restart the pretraining of a thoroughly trained LLM?

% 

%% file: sections/texonomy.tex
\tikzstyle{my-box}=[
    rectangle,
    draw=hidden-draw,
    rounded corners,
    text opacity=1,
    minimum height=1.5em,
    minimum width=5em,
    inner sep=2pt,
    align=center,
    fill opacity=.5,
    line width=0.8pt,
]
\tikzstyle{leaf}=[my-box, minimum height=1.5em,
    text=black, align=left,font=\normalsize,
    inner xsep=2pt,
    inner ysep=4pt,
    line width=0.8pt,
]
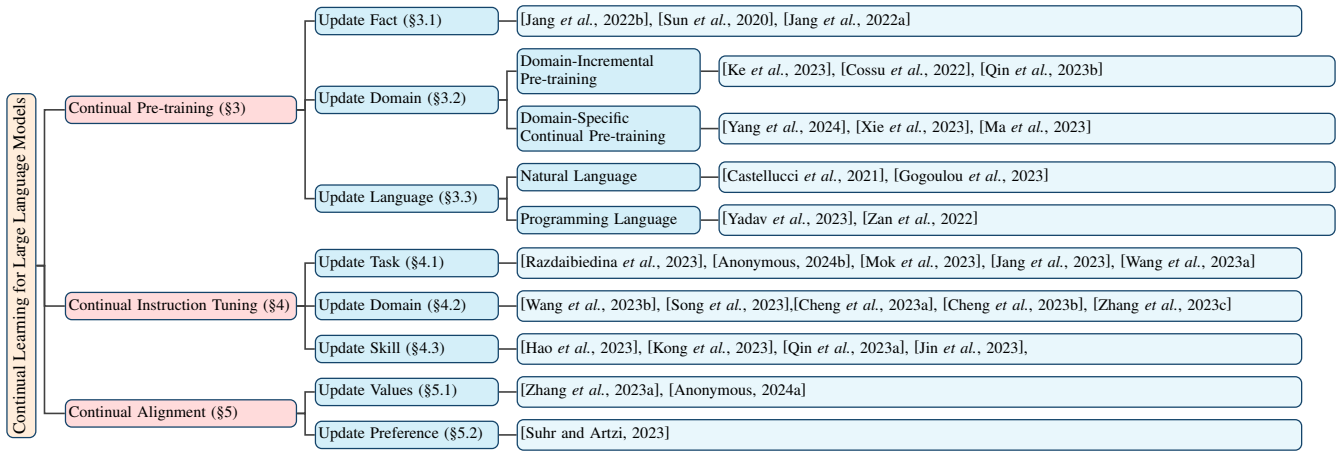
\begin{figure*}[th!]
    \centering
    \resizebox{\textwidth}{!}{
        \begin{forest}
            forked edges,
            for tree={
                grow=east,
                reversed=true,
                anchor=base west,
                parent anchor=east,
                child anchor=west,
                base=left,
                font=\large,
                rectangle,
                draw=hidden-draw,
                rounded corners,
                align=left,
                minimum width=4em,
                edge+={darkgray, line width=1pt},
                s sep=7pt,
                inner xsep=2pt,
                inner ysep=3pt,
                line width=0.8pt,
                ver/.style={rotate=90, child anchor=north, parent anchor=south, anchor=center},
            },
            where level=1{text width=14em,font=\normalsize,}{},
            where level=2{text width=11em,font=\normalsize,}{},
            where level=3{text width=11em,font=\normalsize,}{},
            where level=4{text width=10em,font=\normalsize,}{},
            [
                Continual Learning for Large Language Models,fill=hidden-yellow!70,ver
                [
                    Continual Pre-training (\S \ref{sect:cpt}),fill=hidden-red!70
                    [
                     Update Fact  (\S \ref{sect:cpt_fact}),fill=hidden-blue!70
                            [
                                \cite{JangYYSHKCS22}{, }\cite{SunWLFTWW20}{, }\cite{JangYLYSHKS22}, leaf, text width=48.5em,fill=hidden-blue!70
                            ]
                    ]
                    [
                     Update Domain  (\S \ref{sect:cpt_domain}),fill=hidden-blue!70
                        [
                            Domain-Incremental \\ Pre-training,fill=hidden-blue!70
                            [
                                \cite{KeSLKK023}{, }\cite{abs-2205-09357}{, }\cite{QinQHLWXLSZ23}
                                , leaf, text width=38em,fill=hidden-blue!70
                            ]
                        ]
                        [
                            Domain-Specific \\ Continual Pre-training,fill=hidden-blue!70
                            [
                                \cite{abs-2401-01600}{, }\cite{abs-2311-08545}{, }\cite{abs-2312-15696}
                                % {, }\cite{abs-2311-12315}
                                    , leaf, text width=38em,fill=hidden-blue!70
                            ]
                        ]
                    ]
                    [
                     Update Language  (\S \ref{sect:cpt_language}),fill=hidden-blue!70
                        [
                            Natural Language,fill=hidden-blue!70
                            [
                                \cite{CastellucciFC020}{, }\cite{abs-2311-01200}
                                , leaf, text width=38em,fill=hidden-blue!70
                            ]
                        ]
                        [
                            Programming Language,fill=hidden-blue!70
                            [
                                \cite{YadavSDLZTBMNRB23}{, }\cite{ZanCYLKGWCL22}, leaf, text width=38em,fill=hidden-blue!70
                            ]
                        ]
                    ]
                ]
                [
                    Continual Instruction Tuning (\S \ref{sect:cit}),fill=hidden-red!70
                    [
                        Update Task  (\S \ref{sect:cit_task}),fill=hidden-blue!70
                            [
                                \cite{razdaibiedina2023progressive}{, }\cite{anonymous2024scalable}{, }\cite{mok2023large}{, }\cite{jang2023exploring}{, }\cite{wang2023orthogonal}
                                , leaf, text width=48.5em,fill=hidden-blue!70
                            ]
                    ]
                    [
                        Update Domain  (\S \ref{sect:cit_domain}),fill=hidden-blue!70
                            [
                                \cite{wang2023trace}{, }\cite{song2023conpet}{,}\cite{cheng2023adapting}{, }\cite{cheng2023language}{, }\cite{zhang2023reformulating}
                                % {, }\cite{jin2023genegpt}
                                , leaf, text width=48.5em,fill=hidden-blue!70
                            ]
                    ]
                    [
                        Update Skill  (\S \ref{sect:cit_tool}),fill=hidden-blue!70
                            [
                                \cite{hao2023toolkengpt}{, }\cite{kong2023tptu}{, }\cite{qin2023toolllm}{, }\cite{jin2023genegpt}{, }
                                , leaf, text width=48.5em,fill=hidden-blue!70
                            ]
                    ]
                ]
                [
                    Continual Alignment (\S \ref{sect:ca}),fill=hidden-red!70
                    [
                        Update Values  (\S \ref{sect:ca_value}),fill=hidden-blue!70
                            [
                                \cite{zhang2023copf}{, }\cite{anonymous2023cppo}
                                , leaf, text width=48.5em,fill=hidden-blue!70
                            ]
                    ]
                    [
                        Update Preference  (\S \ref{sect:ca_preference}),fill=hidden-blue!70
                            [
                                \cite{suhr2023continual}
                                , leaf, text width=48.5em,fill=hidden-blue!70
                            ]
                    ]
                ]
            ]
        \end{forest}
    }
    % \vspace{.3cm}
    \caption{Taxonomy of trends in continual learning for large language models.}
    \label{fig:survey}
\end{figure*}

%% file: sections/3_Pretraining.tex
% Please add the following required packages to your document preamble:
% \usepackage{booktabs}
\begin{table}[htb]
\centering
\resizebox{\columnwidth}{!}{
\begin{tabular}{@{}cccc@{}}
\toprule
Information                  & Pretraining & Instruction-tuning & Alignment \\ \midrule
Fact         & \faCheckCircleO           & ×                  & ×          \\
Domain            & \faCheckCircleO           & \faCheckCircleO                  & ×          \\
Language          & \faCheckCircleO           & ×                  & ×          \\
Task              & ×           & \faCheckCircleO                  & ×          \\
Skill (Tool use) & ×           & \faCheckCircleO                  & ×          \\
Value            & ×           & ×                  & \faCheckCircleO          \\
Preference       & ×           & ×                  & \faCheckCircleO          \\ \bottomrule
\end{tabular}}
\caption{Information updated during different stages of continual learning for LLMs.}\label{tab:information}
\end{table}

\section{Continual Pre-training (CPT)}
\label{sect:cpt}
Continual pretraining in large language models is essential for keeping the LLMs relevant and effective. This process involves regularly updating the models with the latest information~\cite{JangYLYSHKS22}, adapting them to specialized domains~\cite{KeSLKK023}, enhancing their coding capabilities~\cite{YadavSDLZTBMNRB23}, and expanding their linguistic range~\cite{CastellucciFC020}. With CPT, LLMs can stay current with new developments, adapt to evolving user needs, and remain effective across diverse applications. Continual pretraining ensures LLMs are not just knowledgeable but also adaptable and responsive to the changing world.

\subsection{CPT for Updating Facts}
\label{sect:cpt_fact}
% Integrating Recent Information
The capability of LLMs to integrate and adapt to recent information is crucial. A pivotal strategy here is the employment of dynamic datasets that facilitate the real-time assimilation of data from a variety of sources like news feeds~\cite{SunWLFTWW20}, scholarly articles~\cite{abs-2205-09357}, and social media~\cite{abs-2205-09357}. 
% % Handling Evolving Facts and Concepts
% Addressing the challenge of evolving facts~\cite{JangYLYSHKS22} and concepts~\cite{OnoeZPDC23}, researchers have been exploring the utility of information versioning systems. These systems are designed to keep track of the historical changes in knowledge. 
\cite{SunWLFTWW20} presents ERNIE 2.0, which is a continual pre-training framework that incrementally builds and learns from multiple tasks to maximize knowledge extraction from training data. \cite{JangYYSHKCS22} introduces continual knowledge learning, a method for updating temporal knowledge in LLMs, reducing forgetting while acquiring new information. \cite{JangYLYSHKS22} shows that continual learning with different data achieves comparable or better perplexity in language models than training on the entire snapshot, confirming that factual knowledge in LMs can be updated efficiently with minimal training data. Integral to this process is the implementation of automated systems for the verification of newly acquired data, ensuring both the accuracy and dependability of the information. 

\subsection{CPT for Updating Domains}
\label{sect:cpt_domain}
Continual pre-training updates domain knowledge through two approaches: 1) domain-incremental pre-training accumulates knowledge across multiple domains, and 2) domain-specific continual learning, which evolves a general model into a domain expert by training on domain-specific datasets and tasks.
%  Domain-incremental learning
% 
In domain-incremental pre-training, \cite{abs-2205-09357} explores how models can be continually pre-trained on new data streams for both language and vision, preparing them for various downstream tasks. \cite{QinQHLWXLSZ23} examines continual retraining by assessing model compatibility and benefits of recyclable tuning via parameter initialization and knowledge distillation. \cite{KeSLKK023} introduces a soft-masking mechanism to update language models (LMs) with domain corpora, aiming to boost performance while preserving general knowledge.
% 
% Domain-specific continual learning
For domain-specific continual learning, \cite{abs-2311-08545} develops FinPythia-6.9B through domain-adaptive pre-training for the financial sector. EcomGPT-CT~\cite{abs-2312-15696} investigates the effects of continual pre-training in the E-commerce domain. 
%AcademicGPT~\cite{abs-2311-12315} focuses on academic content, utilizing a corpus of academic papers and high-quality Chinese data, derived from LLaMA2-70B. 
These studies collectively highlight the evolving landscape of continual pre-training, demonstrating its effectiveness in enhancing model adaptability and expertise across a wide range of domains.

\subsection{CPT for Language Expansion}
\label{sect:cpt_language}
% Adding New Languages
Expanding the range of languages that LLMs can understand and process is essential for ensuring broader accessibility~\cite{CastellucciFC020}. This expansion is not just about including a wider variety of languages, particularly underrepresented ones, but also about embedding cultural contexts into language processing. 
% Enhancing Understanding of Dialects and Slang
A significant challenge here is the model's ability to recognize and interpret regional dialects and contemporary slangs~\cite{abs-2311-01200}, which is crucial for effective and relevant communication across diverse racial, social and cultural groups.

% Incorporating New Programming Languages
In addition to mastering natural languages, LLMs have also made significant strides in understanding and generating programming languages. 
\cite{YadavSDLZTBMNRB23}  introduced CodeTask-CL, a benchmark for continual code learning that encompasses a diverse array of tasks, featuring various input and output formats across different programming languages. 
\cite{ZanCYLKGWCL22} explore using an unlabeled code corpus for training models on library-oriented code generation, addressing the challenge of scarce text-code pairs due to extensive library reuse by programmers. They introduce CERT, a method where a "sketcher" outlines a code structure, and a "generator" completes it, both continuously pre-trained on unlabeled data to capture common patterns in library-focused code snippets. These developments highlight LLMs' potential to transform both natural and programming language processing, leading to more efficient coding practices.

%% file: sections/4_Instruction.tex
\section{Continual Instruction Tuning (CIT)}
\label{sect:cit}
% Continual Pertaining finetunes LLMs on a stream of incoming corpus in an unsupervised manner to learn new knowledge. 
LLMs have shown great instruction following abilities that can be used to complete different tasks with a few-shot task prompt. Continual Instruction Tuning (CIT) involves continually finetuning the LLMs to learn how to follow instructions and transfer knowledge for future tasks \cite{zhang2023citb}. Based on the ability and knowledge updated during instruction tuning, we can further divide CIT into three categories: \emph{1) task-incremental CIT}, \emph{2) domain-incremental CIT}, and \emph{tool-incremental CIT}.

\subsection{Task-incremental CIT}
\label{sect:cit_task}
% \subsection{Task-Specific Tuning}
Task-incremental Continual Instruction Tuning (Task-incremental CIT) aims to continuously finetune LLMs on a sequence of task-specific instructions and acquire the ability to solve novel tasks.  A straightforward solution is to continuously generate instruction-tuning data for new tasks and directly fine-tune LLMs on it \cite{wang2023trace}. 
% catastrophic forgetting
However, studies have shown that continuously finetuning LLMs on task-specific data would cause a catastrophic forgetting of the learned knowledge and problem-solving skills in previous tasks \cite{kotha2023understanding}. TAPT \cite{gururangan2020don} presents a simple data selection strategy that retrieves unlabeled text from the in-domain corpus, aligning it with the task distribution. This retrieved text is then utilized to fine-tune LLMs, preventing catastrophic forgetting and enhancing argument performance. To mitigate catastrophic forgetting, Contunual-T0 \cite{scialom2022fine} employs rehearsal with a memory buffer \cite{shin2017continual} to store previous tasks data and replay them during training. ConTinTin \cite{yin2022contintin} presents InstructionSpeak, which includes two strategies that make full use of task instructions to improve forward-transfer and backward-transfer. The first strategy involves learning from negative outputs, while the second strategy focuses on revisiting instructions from previous tasks. RationaleCL \cite{xiong2023rationale} conducts contrastive rationale replay to alleviate catastrophic forgetting. DynaInst \cite{mok2023large} proposes a hybrid approach incorporating a Dynamic Instruction Replay and a local minima-inducing regularizer. These two components enhance the generalizability of LLMs and decrease memory and computation usage in the replay module.
Unlike previous replay-based or regularization-based methods, SLM \cite{anonymous2024scalable} incorporates vector space retrieval into the language model, which aids in achieving scalable knowledge expansion and management. This enables LLMs' quick adaptation to novel tasks without compromising performance caused by catastrophic forgetting.

% Efficiency
LLMs with billions of parameters introduce a huge computational burden for conducting continual learning. To address this issue, the Progressive Prompts technique \cite{razdaibiedina2023progressive} freezes the majority of parameters and only learns a fixed number of tokens (prompts) for each new task. Progressive Prompts significantly reduce the computational cost while alleviating catastrophic forgetting and improving the transfer of knowledge to future tasks. ELM \cite{jang2023exploring} first trains a small expert adapter on top of the LLM for each task. Then, it employs a retrieval-based approach to choose the most pertinent expert LLM for every new task. Based on the parameter-efficient tuning (PET) framework,
% (e.g., LoRA \cite{hu2021lora}), 
O-LoRA \cite{wang2023orthogonal} proposes an orthogonal low-rank adaptation for CIT. O-LoRA incrementally learns new tasks in an orthogonal subspace while fixing the LoRA parameters learned from past tasks to minimize catastrophic forgetting. Similarly, DAPT \cite{zhao2024dapt} proposes a novel Dual Attention Framework to align the learning and selection of LoRA parameters via the Dual Attentive Learning\&Selection module. LLaMA PRO \cite{wu2024llama} proposes a novel block expansion technique, which enables the injection of new knowledge into LLMs and preserves the initial capabilities with efficient post-training. 

\subsection{Domain-incremental CIT}
\label{sect:cit_domain}
% \subsection{Domain Adaptation}
Domain-incremental Continual Instruction Tuning (Domain-incremental CIT) aims to continually finetune LLMs on a sequence of domain-specific instructions and acquire the knowledge to solve tasks in novel domains. TAPT \cite{gururangan2020don} adaptively tunes the LLMs on a series of domain-specific data including biomedicine, computer science, news, and shopping reviews. Then, it evaluates the LLMs' text classification ability in each domain. ConPET \cite{song2023conpet} applies previous continual learning methods, initially developed for smaller models, to LLMs using PET and a dynamic replay strategy. This approach significantly reduces tuning costs and mitigates overfitting and forgetting problems. Experiments conducted on a typical continual learning scenario, where new knowledge types gradually emerge, demonstrate the superior performance of ConPET. AdaptLLM \cite{cheng2023adapting} adapts LLMs to different domains by enriching the raw training corpus into a series of reading comprehension tasks relevant to its content. These tasks are designed to help the model leverage domain-specific knowledge while enhancing prompting performance. PlugLM \cite{cheng2023language} uses a differentiable plug-in memory (DPM) to explicitly store the domain knowledge. PlugLM could be easily adapted to different domains by plugging in in-domain memory. \cite{zhang2023reformulating} designs an adapt-retrieve-revise process that adapts LLMs to new domains. It first uses the initial LLMs' respose to retrieve knowledge from the domain database. The retrieved knowledge is used to revise initial responses and obtain final answers. \cite{dong2023abilities} analyze the LLMs continuously tuned on different domains and find that the sequence of training data has a significant impact on the performance of LLMs. They also offer a Mixed Fine-tuning (DMT) strategy to learn multiple abilities in different domains.

\subsection{Tool-incremental CIT}
\label{sect:cit_tool}
% \subsection{Tool Integration}
Tool-incremental Continual Instruction Tuning (Tool-incremental CIT) aims to fine-tune LLMs continuously, enabling them to interact with the real world and enhance their abilities by integrating with tools, such as calculators, search engines, and databases \cite{qin2023toolllm}. With the rapid emergence of new tools like advanced software libraries, novel APIs, or domain-specific utilities \cite{liang2023taskmatrix,jin2023genegpt}, there is a growing need to continually update LLMs so they can quickly adapt and master these new tools. Llemma \cite{azerbayev2023llemma} continues tuning LLMs on a dataset with a mixture of math-related text and code to enable LLMs to solve mathematical problems by using external tools. ToolkenGPT \cite{hao2023toolkengpt} represents each tool as a new token (toolken) whose embedding is learned during instruction tuning. This approach offers an efficient way for LLMs to master tools and swiftly adapt to new tools by adding additional tokens. 

%% file: sections/5_Preference.tex
\section{Continual Alignment (CA)}
\label{sect:ca}

LLMs need to adapt to evolving societal values, social norms and ethical guidelines. Furthermore, there exists substantial diversity in preferences across different demographic groups, as well as individuals' changing preferences over time. The need to respond to these changes give rise to continual alignment. 
In the context of continual alignment, two scenarios emerge: (i) the requirement to update LLMs to reflect shifts in societal values
% , which involves tasks related to model editing and unlearning outdated values, 
and (ii) integrating new demographic groups or value types to existing LLMs, which we will describe in the following subsections.

\subsection{Continual Value Alignment} 
\label{sect:ca_value}
Continual value alignment aims to continually incorporate ethical guidelines or adapt to cultural sensitivities and norms.
% , a problem aligned with continual learning. 
It requires updating to unlearn outdated notions and incorporating new values, akin to model editing and unlearning tasks. Model editing and knowledge unlearning have been studied in pretraining and instruction tuning phases~\cite{yao-etal-2023-editing}; however, they have not yet been explored in preference learning. 

\subsection{Continual Preference Alignment} 
\label{sect:ca_preference}
Adding new demographic groups or value types aligns with continual learning problems, aiming to guide LLMs in generating responses aligned with emerging values while adhering to previously learned ones. For example, many open-source aligned LLMs employ reinforcement learning with human feedback (RLHF) for safety. We may want to align the LLMs for additional attributes such as helpfulness and faithfulness. Beyond the challenge of retaining past preferences while maximising the reward on new ones, continual preference learning also faces difficulties in stable and efficient training with a large action space (vocabulary) and a large number of parameters. 
Previous works have demonstrated proof-of-concept of such agents. However, there is a lack of standardized benchmarks to systematically evaluate the learning capabilities of new preferences over time. 
Continual Proximal Policy Optimization (CPPO)~\cite{anonymous2023cppo} utilizes a sample-wise weighting on the Proximal Policy Optimization (PPO) algorithm~\cite{schulman2017proximal} to balance policy learning and knowledge retention in imitating the old policy output. On the other hand, \cite{zhang2023copf} extend the Direct Preference Optimization (DPO) algorithm \cite{rafailov2023direct} to the continual learning setting by employing Monte Carlo estimation to derive a sequence of optimal policies for the given sequences of tasks and incorporate them to regularize the policy learning on new tasks. 
%Due to the lack of continual preference benchmarks, these previous works provide are proof of concept only.

% Challenges: no dataset, benchmark

% - Tailoring Responses to Individual User Preferences
% - Learning from User Feedback and Interactions

% \cite{suhr2023continual} designs a contextual bandit learning approach where human users instruct an agent using natural language.

% \subsection{Why?}
% \begin{itemize}
%     \item Change in values and preferences~\cite{gabriel_artificial_2020}: Human and societal values change over time, but less frequently. On the other hand, preferences tend to be easily influenced and change more often.
%     \item Preference diversity: each individual and demographic group has its own taste and requirement.
% \end{itemize}

% Therefore, there is a need to update the aligned-LLMs to reflect the changes in values and preferences as well as to align the LLMs with new demographic group.

% \subsection{Problem Formulation}

% \subsection{How?}

%% file: sections/6_Evaluation.tex
\section{Benchmarks}
\label{sect:eval}
The systematic evaluation of LLMs' continual learning performance demands benchmarks with high-quality data sources and diverse content. Below we summarize notable benchmark dataets. 

\subsection{Benchmarks for CPT}
TemporalWiki~\cite{JangYLYSHKS22} serves as a lifelong benchmark, training and evaluating Language Models using consecutive snapshots of Wikipedia and Wikidata, helping assess an LM's ability to retain past knowledge and acquire new knowledge over time. Additional social media datasets like Firehose~\cite{HuSSK23} comprise 100 million tweets from one million users over six years. CKL~\cite{JangYYSHKCS22} focuses on web and news data, aiming to retain time-invariant world knowledge from initial pretraining while efficiently learning new knowledge through continued pre-training on different corpora. TRACE~\cite{wang2023trace} encompasses eight diverse datasets covering specialized domains, multilingual tasks, code generation, and mathematical reasoning. These datasets are harmonized into a standard format, facilitating straightforward and automated evaluation of LLMs. Due to the fast-paced nature of data, time-sensitive datasets quickly become outdated, necessitating frequent updates to continual pre-training benchmarks for model evaluation.

\subsection{Benchmarks for CIT} The Continual Instruction Tuning Benchmark (CITB)~\cite{zhang2023citb} is based on SuperNI, encompassing over 1,600 Natural Language Processing (NLP) tasks across 76 types like language generation and classification, all in a text-to-text format. ConTinTin~\cite{yin2022contintin}, another benchmark derived from NATURAL-INSTRUCTIONS, includes 61 tasks across six categories, such as question generation and classification.
When using these benchmarks for evaluating black-box language learning models that cannot access their training data, the selection of datasets is crucial to avoid task contamination and ensure reliable performance assessment in continual instruction tuning.

\subsection{Benmarks for CA}
COPF~\cite{zhang2023copf} conduct experiments for continual alignment using datasets like the Stanford Human Preferences (SHP)~\cite{EthayarajhCS22} and Helpful \& Harmless (HH) Datasets~\cite{abs-2204-05862}. The SHP Dataset comprises 385,000 human preferences across 18 subjects, from cooking to legal advice. The HH Dataset consists of two parts: one where crowdworkers interact with AI models for helpful responses, and another where they elicit harmful responses, selecting the more impactful response in each case. 
% \subsection{Metric}
Despite the growing interest in this field, there is a notable absence of dedicated benchmarks for continual alignment, presenting an opportunity for future research and development in this area.

\section{Evaluation}
\subsection{Evaluation for Target Task Sequence}
Continual learning for large language models involves evaluating the model's performance over a task sequence. Performance can be measured by three typical continual learning metrics: (1) average performance; (2) Forward Transfer Rate (FWT), and (3) Backward Transfer Rate (BWT)~\cite{lopez2017gradient,WuCLLQH22}: 

(1) FWT assesses the impact of knowledge acquired from previous tasks on the initial ability to perform a new task, prior to any dedicated training for that new task. 
\begin{align}
    FWT = \frac{1}{T-1}\sum_{i=2}^{T-1}A_{T,i} - \tilde{b_{i}}
\end{align}

(2) BWT measures catastrophic forgetting by comparing a model's performance on old tasks before and after learning new ones.
\begin{align}
    BWT = \frac{1}{T-1}\sum_{i=1}^{T-1}A_{T,i} - A_{i,i}
\end{align}

(3) Average Performance, e.g., the average accuracy assesses the ability of a model or algorithm to effectively learn from and adapt to a sequence of data streams or tasks over time.
\begin{align}
    Avg.\ ACC = \frac{1}{T}\sum_{i=1}^{T}A_{T,i}
\end{align}
where $A_{t,i}$ is the accuracy of models on the test set of $i$th task after model learning on the $t$th task and $\tilde{b_{i}}$ is the test accuracy for task $i$ at random initialization. 
% In addition to FWT and BWT, we  consider average accuracy as a key performance measure, measuring the accuracy of past tasks after the model has moved on to learning new tasks.

\subsection{Evaluation for Cross-stage Forgetting}
Large language models continually trained on different stages can experience the issue of unconscious forgetting \cite{abs-2309-06256}, which shows that continual instruction tuning can erode the LLM's general knowledge. Additionally, previous studies \cite{abs-2310-03693} also demonstrate that the behavior of safely aligned LLMs can be easily affected and degraded by instruction tuning. To quantify these limitations, TRACE~\cite{wang2023trace} proposes to evaluate LLMs by using three novel metrics: General Ability Delta (GAD), Instruction Following Delta (IFD), and Safety Delta (SD):

(1) GAD assesses the performance difference of an LLM on general tasks after training on sequential target tasks.
\begin{align}
    GAD = \frac{1}{T} \sum_{i=1}^T ( R_{t, i}^G - R_{0, i}^G)
\end{align}

(2) IFD assesses the changes of model's instruction-following ability after training on sequential different tasks.
\begin{align}
    IFD = \frac{1}{T} \sum_{i=1}^T (R_{t, i}^I - R_{0, i}^I)
\end{align}

(3) SD assesses the safety variation of a model's response after sequential training.
\begin{align}
    SD = \frac{1}{T} \sum_{i=1}^T (R_{t, i}^S - R_{0, i}^S)
\end{align}
The baseline performance of the initial LLM on the $i$-th task is represented by $ R_{0,i} $. After incrementally learning up to the $t$-th task, the score on the $i$-th task becomes $ R_{t,i}$. And $R^G$, $R^I$, and $R^S$ represent the performance of LLM on general tasks (assessing the information obtained from pre-training), instruction-following tasks, and alignment tasks, respectively.
These measure changes in an LLM's overall capabilities, adherence to instructions, and safety after continual learning, going beyond traditional benchmarks by focusing on maintaining inherent skills and aligning with human preferences.

%% file: sections/7_Challenges_and_Future_Directions.tex
\section{Challenges and Future Works}

\paragraph{Computation-efficient Continual Learning}

In the realm of computation efficiency, the focus is on enhancing the continual pretraining process with minimized computational resources~\cite{abs-2311-11908}. This involves developing innovative architectures that can handle the increasing complexity of pretraining tasks without proportional increases in computational demands. Efficiency in algorithms and data structures becomes crucial, especially in managing the extensive data involved in pretraining. Additionally, energy-efficient learning models are vital for sustainable scaling of LLMs, aligning with Green AI initiatives. This area requires balancing the computational cost vs the benefits in terms of model performance and capabilities.

\paragraph{Social Good Continual Learning}

Social responsibility in continual learning encompasses ensuring privacy and data security, particularly in the context of continual instruction tuning~\cite{gabriel_artificial_2020}. As LLMs are fine-tuned with more specific instructions or tasks, the handling of sensitive or personal data must be managed securely and ethically. Aligning with human values and culture is also paramount, especially in the realm of continual preference learning. This involves incorporating ethical AI principles and cultural sensitivities to ensure that the model's outputs are aligned with societal norms and values.

\paragraph{Automatic Continual Learning}
A significant challenge lies in creating systems capable of autonomously overseeing their learning processes, seamlessly adjusting to novel tasks (instruction tuning) and user preferences (alignment) while relying solely on the inherent capabilities of LLMs, all without the need for manual intervention~\cite{qiao2024autoact}. Automatic continual learning includes multi-agent systems capable of collaborative learning and self-planning algorithms that can autonomously adjust learning strategies based on performance feedback. Such systems would represent a significant advancement in the autonomy of LLMs.

\paragraph{Continual Learning with Controllable Forgetting}
Controllable forgetting is particularly relevant to continual pretraining. The ability to selectively retain or forget information as the model is exposed to new data streams can prevent catastrophic forgetting~\cite{abs-2310-03693} and enhance model adaptability~\cite{wang2023trace}. This challenge also extends to managing misinformation and unlearning incorrect or outdated information~\cite{ChenY23}, ensuring the accuracy and reliability of the LLM over time.

\paragraph{Continual Learning with History Tracking}
Effective history tracking is vital for understanding the evolution of the LLM through its phases of pre-training, instruction tuning, and preference learning. Managing history in model parameters and using external memory architectures can help in tracking the influence of past learning on current model behavior and decisions~\cite{abs-2302-07842}. This is crucial for analyzing the effectiveness of continual learning processes and making informed adjustments.

\paragraph{Theoretical insights on  LLM in Continual Learning. }
Numerous evaluation studies have examined the issue of cross-stage forgetting~\cite{abs-2309-06256} and demonstrated the weak robustness of aligned LLMs~\cite{abs-2310-03693}. However, theoretical analyses of how multi-stage training impacts the performance of large language models in subsequent continual learning tasks are scarce. This gap highlights the need for a deeper understanding of the specific changes multi-stage training introduces to LLMs' learning capabilities and long-term performance.

%% file: sections/8_Conclusion.tex
\section{Conclusion}
Continual learning holds the vital importance of allowing large language models to be regularly and efficiently updated to remain up-to-date with the constantly changing human knowledge, language and values. 
% We differentiate continual learning from other enhancement strategies such as retrieval-augmented generation and model editing, focus specifically on improving LLMs' linguistic and cognitive abilities. 
We showcase the complex, multi-stage process of continual learning in LLMs, encompassing continual pretraining, instruction tuning, and alignment, a paradigm more intricate than those used in continual learning on smaller models. As the first survey of its kind to thoroughly explore continual learning in LLMs, this paper categorizes the updates by learning stages and information types, providing a detailed understanding of how to effectively implement continual learning in LLMs. %, and contributing to the development of more advanced and adaptable language models.
With a discussion of major challenges and future work directions, our goal is to provide a comprehensive account of recent developments in continual learning for LLMs, shedding light on the development of more advanced and adaptable language models.